\documentclass[%
 reprint,
 amsmath,amssymb,
 aps,
]{revtex4-2}

\usepackage{graphicx}% Include figure files
\usepackage{dcolumn}% Align table columns on decimal point
\usepackage{bm}% bold math
\begin{document}

% \preprint{APS/123-QED}

\title{Isotropic Fourier Neural Operators}

\author{Michael F. Staddon}
%\email{staddon@mpi-cbg.de}
\affiliation{Independent Researcher}
% \affiliation{Max Planck Institute of Molecular Cell Biology and Genetics, Dresden, Germany}
% \affiliation{Center for Systems Biology Dresden, Dresden, Germany}
% \affiliation{Cluster of Excellence, Physics of Life, TU Dresden, Dresden, Germany}

\begin{abstract}

% Introduction
Fourier Neural Operators are deep learning models that learn mappings between function spaces and can be used to learn and solve partial differential equations (PDEs), in some cases significantly faster than traditional PDE solvers. Within the model are Fourier layers, which apply linear transformations directly to the Fourier modes, with parameters depending on the wave numbers. However, most physical systems are isotropic, with the results being independent of the coordinate system chosen, but the linear transformations do not necessarily respect these symmetries. We propose a modification to the linear transformations that ensures spatial symmetries are respected, called the Isotropic Fourier Neural Operator, which both improves model performance and reduces the number of parameters by up to a factor of 16 in 2D and 96 in 3D.

\end{abstract}
\maketitle

\section{Introduction}

% TO DO: Check all comments
% - Plot simulation predictions! Eg Input, Actual, FNO, ISO + Rotated

Partial differential equations (PDEs) are used to model a range of problems across the sciences, engineering, and economics, such as fluid dynamics~\cite{temam2024navier} and weather forecasting~\cite{kalnay2003atmospheric}, traffic flow~\cite{lighthill1955kinematic, richards1956shock}, animal patterning~\cite{turing1990chemical}, and financial options pricing~\cite{black1973pricing, merton1971theory}. Often these do not have known analytic solutions (the Navier-Stokes problem famously so~\cite{fefferman2006existence}) and must be approximated through numerical solvers. While traditional solvers like finite element and finite difference methods can solve equations to a desired precision~\cite{lapidus1999numerical}, more recently deep learning models like neural operators have been used to significantly speed up computation~\cite{li2020fourier, bonev2023spherical, duruisseaux2025fourier}. Additionally, in cases where even the PDE is unknown, such as complex biological systems, neural operators may learn the dynamics from given data samples.

% Needs improving
Fourier Neural Operators (FNOs)~\cite{li2020fourier} have been particularly effective and widespread for solving PDEs. FNOs learn to map between function spaces by applying linear transformations directly on the Fourier modes rather than working in physical space, allowing for a resolution-invariant model that can adapt to super resolution data during prediction. FNOs and its variants, such as the Spherical Fourier Neural Operator~\cite{bonev2023spherical}, have been successfully applied to a range of problems such as climate modelling~\cite{watt2025ace2}, medical imaging~\cite{dai2023neural}, and nuclear fusion~\cite{gopakumar2024plasma}. Several modifications to the basic architecture have been proposed that focus on improving model accuracy and generalisability, or reducing the number of parameters and improving efficiency.

% Memory
Reducing the number of parameters of the model helps to reduce the memory requirement of models and reduce training and inference time. Modifications like the Factorized Fourier Neural Operators (F-FNOs)~\cite{tran2021factorized} or Decomposed Fourier Neural Operators (D-FNOs)~\cite{li2025d} reduce the model complexity by factorising the parameters, for example, by having parameters only for the Fourier transform for the $x$ frequencies and $y$ frequencies separately, changing the scaling with the number of modes $m$ from $O(m^2)$ to $O(m)$. Alternatively, low rank decompositions can decompose the kernel tensor into the product of several smaller tensors and keeping only the most informative, with the Multigrid Tensorized Fourier Neural Operator capable of reducing parameters by a factor of 150 while maintaining model accuracy~\cite{kossaifi2023multi}.

At the same time, changes to the internal architecture or training of the FNO that incorporate knowledge of the PDEs or space itself can be used to improve model accuracy. Physics-Informed Neural Operators~\cite{li2024physics} use the same approach as Physics Informed Neural Networks~\cite{raissi2019deep, cai2021physics, white2023stabilized}, in which the loss function is composed of the L2 prediction error and a PDE constraint, an L2 error between the models predicted PDEs and the expected known PDEs. At a more fundamental level, one can incorporate symmetries of the system itself into the FNO. In many problems, the system being modelled is isotropic, with the choice of axis being arbitrary. Transforming the input, such as swapping the $x$ and $y$ coordinates, should result in an output with the same transformation. Radial Fourier Neural Operators (R-FNOs)~\cite{shen2021rotation} accomplish this by making the kernel depend only on the magnitude of the frequency, while Group Equivariant Fourier Neural Operators (G-FNOs)~\cite{helwig2023group} use group convolutions within the FNO to preserve rotational and reflectional symmetries of space.

In this paper, we develop an Isotropic Fourier Neural Operator (Iso-FNO) which both improves model performance while significantly reducing the number of parameters. We show how symmetries constrain the parameters within the Fourier kernel that transforms the Fourier modes, and thus can generate the full kernel from a much smaller set of parameters. For 2D systems this reduces the parameters by approximately a factor of 16, while 3D systems it is a factor of 96. At the same time, the model shows improved performance compared to the standard FNO when measured on test data. These symmetries are general and can be combined with other model types for further improvements in memory and performance.

\section{Fourier Neural Operators}

\begin{figure}[h!]
\includegraphics[width=\columnwidth]{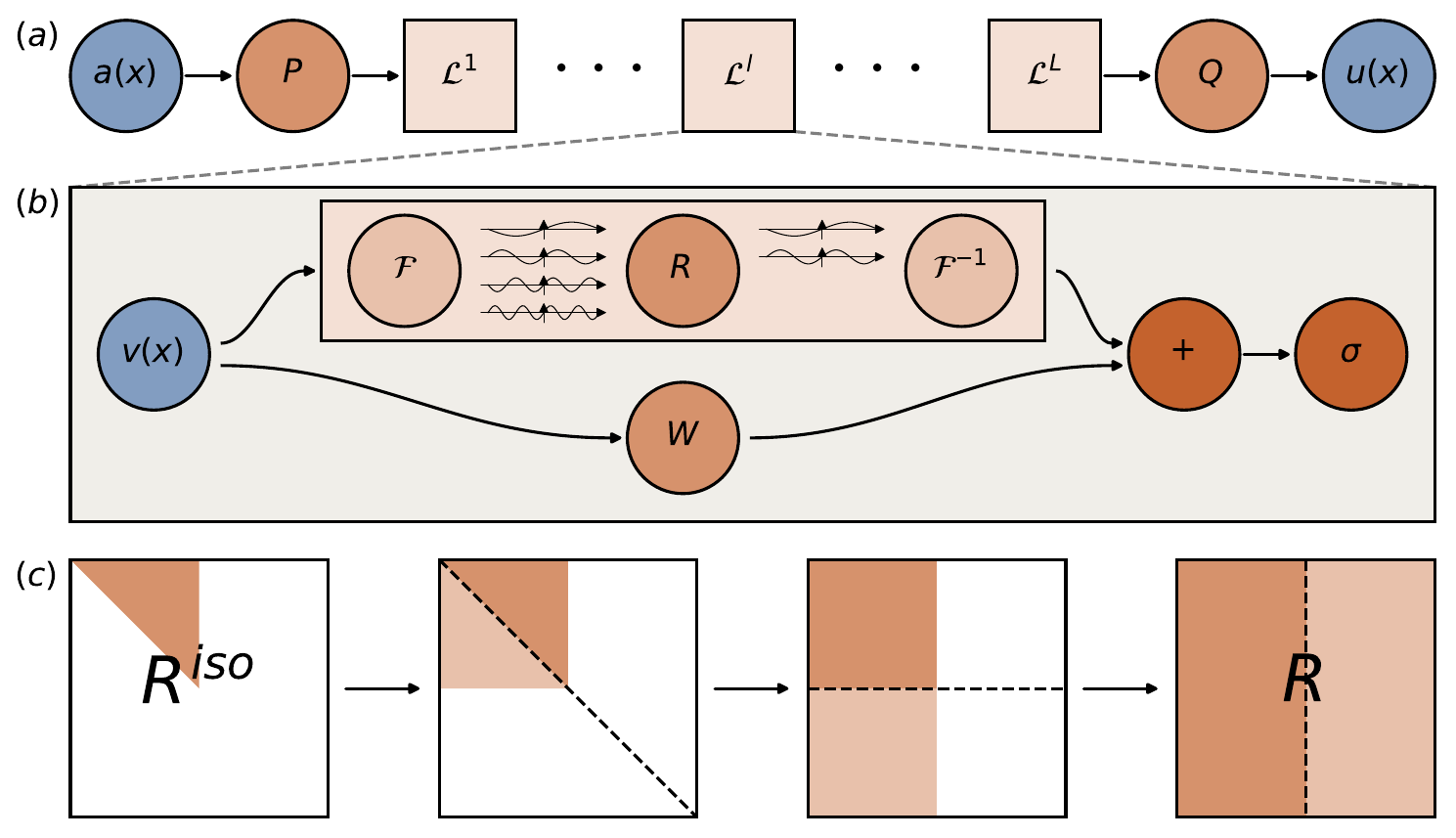}
\caption{Isotropic Fourier Neural Operator. (a) The full model architecture. Lift the input to a higher dimensional channel space by neural network $P$. Apply $L$ Fourier layers. Project down to output channel space by neural network $Q$. (b) Fourier layer architecture. On top: Fourier transform the input, apply linear transform $R$ on the lowest $n$ modes, apply the inverse Fourier transform. On  the bottom: apply linear transform $W$. Finally, add the top and bottom branches and apply non-linear activation function $\sigma$. (c) Isotropic Fourier Kernel. The Fourier kernel $R$ is generated using symmetries of space from an upper right triangular tensor $R^{iso}$. Starting from the left, the original parameters are used to generate the full parameters through reflections representing transposition, reflection in $y$, and reflection in $x$.}
\label{fig:1}
\end{figure}

The original outlining of the Fourier Neural Operator (FNO) uses the following architecture (Fig.~\ref{fig:1}a): given an input function $a(\mathbf{x})$ with $n$ channels lift the input data into a higher dimensional space with tensor $P$, apply several Fourier layers, $L^1 ... L^n$, and project down into the output space with tensor $Q$:
\begin{equation}
   u(\mathbf{x}) = Q \circ \mathcal{L}^L \circ ... \circ \mathcal{L}^2 \circ \mathcal{L}^1  \circ P (a)(\mathbf{x}).
\end{equation}
For example, one might have an input function $a$ with pressure and temperature channels and wish to predict vorticity $v$. In this expression, $P$ and $Q$ are the lifting and projection networks, which are linear transforms acting on the channel dimension of the input. $\mathcal{L}^l$ represents Fourier layer $l$ (Fig.~\ref{fig:1}b) where
\begin{equation}
    \mathcal{L}^l(v)(\mathbf{x}) = \sigma( W^l + \mathcal{F}^{-1} (R^l \cdot \mathcal{F}(v)))(\mathbf{x}),
\end{equation}
where $\sigma$ is a non-linear activation function, in our case ReLU, $W^l$ is a tensor, $\mathcal{F}$ and $\mathcal{F}^{-1}$ are the Fourier and inverse Fourier transforms respectively, and $R^l$ is a tensor that acts linearly on the Fourier modes of the input function $v(\mathbf{x})$. Each complex-valued Fourier kernel $R^l$ only acts on the first $m$ modes, filtering out higher frequency modes. This allows the model to work on different resolutions and can predict super-resolution results from lower resolution training, and for uniform grids the fast Fourier transform is used to speed up the Fourier operations.

The model has hyper-parameters for the number of channels to lift into, $d_v$, the number of Fourier modes to use, $m$, and the number of Fourier layers, $L$, with the majority of parameters being contained in the Fourier kernel when a large number of modes are used. Thus the number of parameters scales with the square of the number of Fourier modes multiplied by the number of channels squared, $O(d_v^2 m^2 L)$. When problems require a higher number of modes to be accurate, the number of parameters in the model increases quadratically which can dramatically increase model size and training time.

\section{Isotropic Fourier Neural Operators}

% Overview of isotropic FNOs eg the symmetries
This paper proposes modifying the Fourier layer to respect common symmetries in 2D space, which we name the Isotropic Fourier Neural Operator (Iso-FNO). These symmetries include reflections in $x$ and $y$, swapping $x$ and $y$ and $90^\circ$ rotations, and whose transformations are described by the dihedral group of order 4, $D_4$. By spatial symmetry, we require that the output of the neural operator transforms in the same way as the input; if we rotate or reflect our input then the corresponding output is also rotated or reflected compared to the original. Since only the Fourier layers involve spatial components, we enforce the parameters of the Fourier kernel $R$ to respect these symmetries, which leads to a 16-fold reduction in the number of parameters in 2 dimensions, and 96-fold in 3D dimensions. 

% Explain transformations
First, consider reflections of the input space in $x$.  Within the Fourier layer, anything transforming a Fourier mode $e^{i (kx + ly)}$ must transform $e^{i(-kx + ly)}$ the same way. Thus we require that (ignoring the channel indices for simplicity) $R_{k,l} = R_{-k,l}$, reducing the number of free parameters by around half. Additionally, if we are Fourier transforming a real valued function and want a real valued output then we must have $R_{k, l} e^{i(kx + ly)} = (R_{-k, l} e^{i(-kx + ly)})^*$ such that the imaginary components cancel out. This implies that $R_{k, l}=R_{-k,l}=R_{-k,l}^*$, meaning all parameters are real valued. Thus we remove all complex values from $R$ compared to the original FNO, again reducing the number of parameters by half. Similar logic may be applied for reflections in $y$, requiring that $R_{k,l} = R_{k,-l}$ and for transpositions, swapping $x$ and $y$, requiring that $R_{k, l} = R_{l, k}$. Each of these requirements reduces the number of free parameters by approximately half, resulting in 1/16th of the number of parameters in the standard FNO, while capturing the isotropy of the system and making the constrained kernel equivariant under the $D_4$ group.

In the isotropic Fourier Neural operator (Iso-FNO), we start with a kernel tensor which is upper diagonal and only has parameters in the left and top halves of the tensor $R^{iso}$, and generate the full Fourier kernel tensor through a series of transpositions and reflections (Fig.~\ref{fig:1}c). This new Fourier kernel ensures $D_4$ group equivariance without needing computationally expensive group convolutions like in the G-FNO~\cite{helwig2023group}. Such a system is generalisable to 3 dimensions, where we expect a 96 factor reduction; an extra 2 from reflections in $z$ and and extra factor of 3 because the permutation group of 3 elements, (our coordinates $x$, $y$, and $z$) has size 6. Model code is available at~\url{https://github.com/mstaddon/iso-fno}

Note that in practice the Isotropic FNO reduces parameters by a factor of around 8 compared to the original implementation of the FNO ~, because the FNO uses the real valued FFT, which only needs the positive modes of one of the dimensions, since the values corresponding to the negative modes must be complex conjugates of the positives mode values as mentioned earlier. Additionally, inference time remains the same for both models since the networks are applying the same tensor calculations with the Fourier modes, where $R$ is generated from the smaller $R^{iso}$ in the Iso-FNO, but the memory size of the model is reduced.

\section{Results}

To test the effectiveness of the Isotropic FNO, we use the 2D Darcy flow which describes the flow of fluid through a porous medium. We model flow within a unit box and no slip boundary conditions:
\begin{align}
     -\nabla \cdot(a(\mathbf{x}) \nabla u(\mathbf{x})) &= f(\mathbf{x}), && \mathbf{x} \in (0, 1)^2 \\
     u(\mathbf{x})&=0, && \mathbf{x} \in \partial(0, 1)^2
\end{align}
where $a(\mathbf{x})$ is the diffusion coefficient, $u(\mathbf{x})$ is the flow speed, and $f(\mathbf{x})$ is the forcing function. The example takes $a(\mathbf{x})$ as the input, with constant forcing $f(\mathbf{x})=1$, and aims to learn the flow $u(\mathbf{x})$.

% Data and training description
For model training and evaluation we use 1000 training instances and 200 testing instances, originally generated by Li et al~\cite{li2020fourier} using a finite difference scheme. We use data with a 128 x 128 resolution. Each model is trained using cosine annealing over 100 epochs with a batch size of 20 and an initial learning rate equal to $0.001$. For each model we use a maximum of 16 modes and 32 channels and aim to minimise the L2 error between predicted and actual flows. As expected, the non-isotropic FNO has 4.2 million parameters while the isotropic FNO has only 0.57 million, a 7.4-fold reduction.

\begin{figure}[h]
\includegraphics[width=\columnwidth]{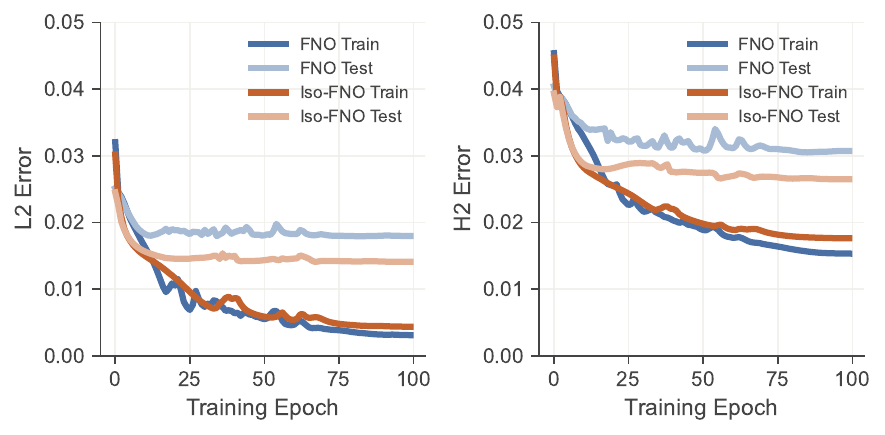}
\caption{L2 error and H2 error for FNO and Iso-FNO against training epoch for the training and test datasets.}
\label{fig:2}
\end{figure}

\begin{figure}[]
\includegraphics[width=\columnwidth]{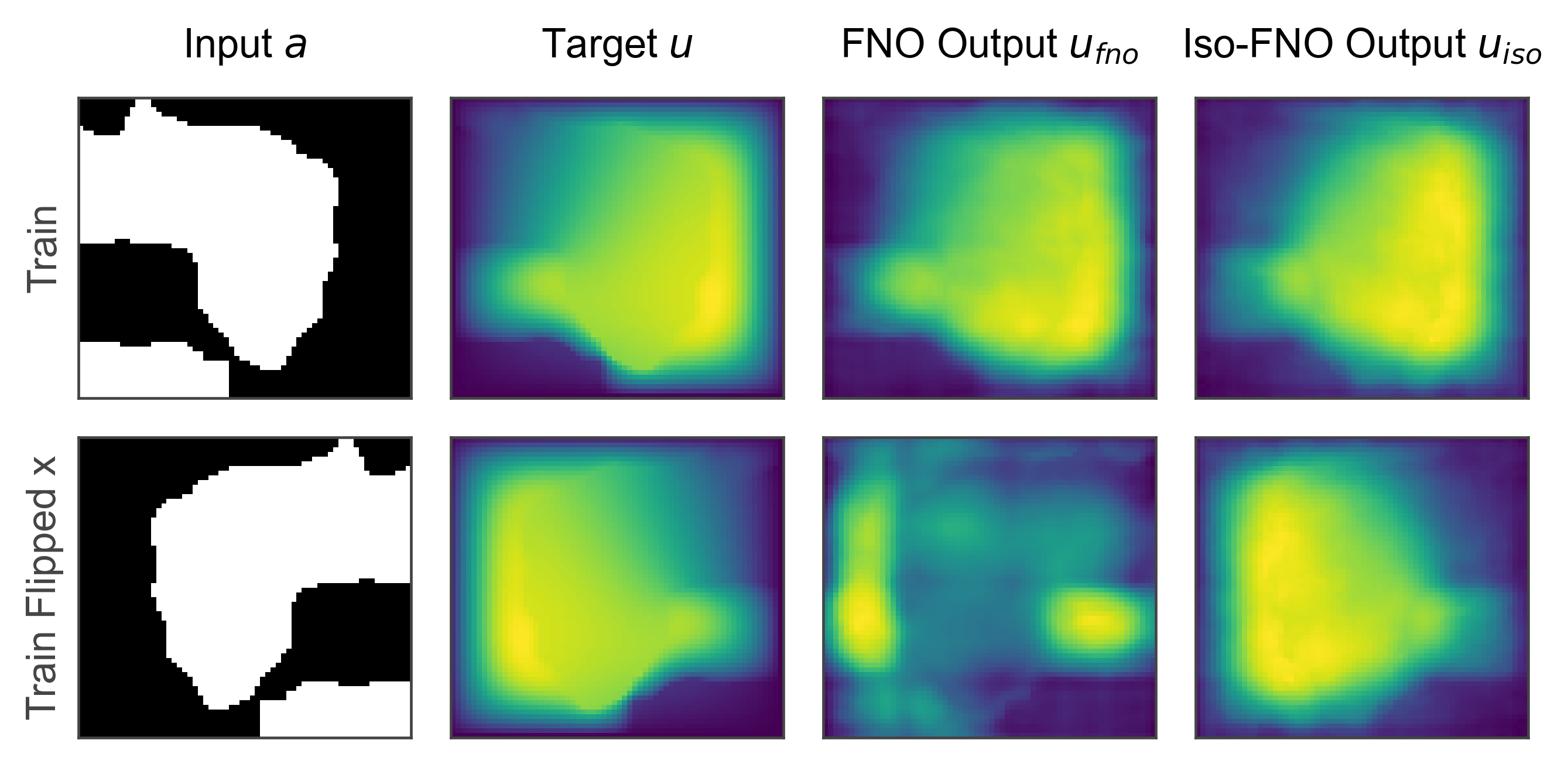}
\caption{FNO and Iso-FNO model predictions on the Darcy Flow problem. Columns: The input function $a$, white has values of 0 and black has values of 1, representing the diffusion coefficient. The ground truth $u$, the flow with no slip boundary conditions, lighter colours are higher values. The FNO prediction $u_{fno}$ and the Iso-FNO prediction $u_{iso}$. Rows: The original data in training, and the data flipped in the x-axis.}
\label{fig:5}
\end{figure}

Our training results are shown in Fig.~\ref{fig:2}. Over the training period, the non-isotropic model has an L2 error about 30\% lower than the isotropic model and 15\% H2 error (Table ~\ref{table:results}). However, performance is reversed on the test data, with the isotropic model having 30\% lower L2 error and 15\% lower H2 error than the non-isotropic model. This is likely because the isotropic FNO has fewer parameters with which to overfit the training data, while also using symmetry to better generalise to out of sample data. From the model predictions (Fig.~\ref{fig:5}) we can see that both FNO and Iso-FNO can predict the target function fairly well. When the input function and target are flipped in the $x$ direction, the FNO's prediction is significantly different from its original prediction flipped and has a high error, while the Iso-FNO is equal to its original prediction flipped.

\begin{table}[h]
    \centering
    \begin{tabular}{r|c|c}
    
         & FNO & Iso-FNO\\
         \hline
       Parameters (m)  &  4.202 & 0.565\\
       \hline
       Train L2  & 0.00309 & 0.00436\\
       Test L2 & 0.01798 & 0.01410\\
       Train H2 & 0.01532 & 0.01763\\
       Test H2 & 0.03072 & 0.02648\\
       \hline
       Transpose L2 & 0.01811 & 0.00436\\
       Flip-x L2 & 0.01778 & 0.00436\\
       Flip-y L2 & 0.01802 & 0.00436\\       
    \end{tabular}
    \caption{Model parameters and errors.}
    \label{table:results}
\end{table}

\begin{figure}[]
\includegraphics[width=\columnwidth]{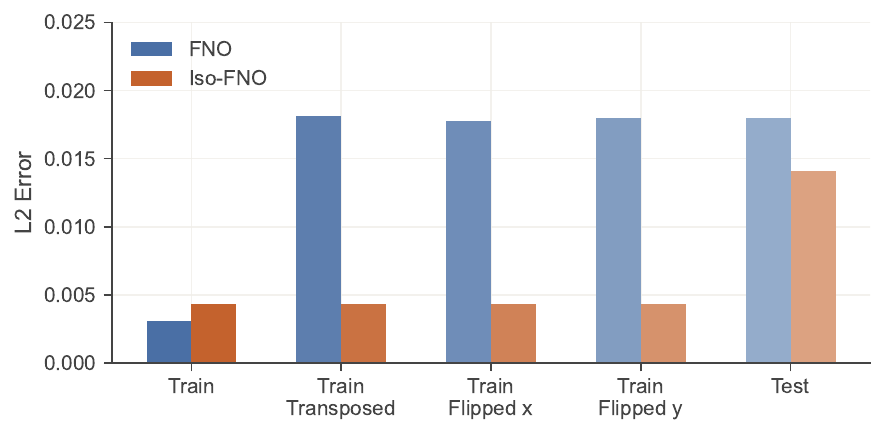}
\caption{L2 error for FNO and Iso-FNO on the training data, the training data transposed or mirrored, and the test data.}
\label{fig:3}
\end{figure}

Additionally, we  measure the error for the two models when using training data but with data flipped in the x and y directions, or swapping x and y. The non-isotropic FNO performs as badly on the transformed training data as on the test data, as shown in Fig.~\ref{fig:5}, while the isotropic FNO generalises to it by the symmetry constraints within the Fourier kernel (Fig.~\ref{fig:3}, Table~\ref{table:results}).

% Look at inference or training speed of IsoFNO with and without parameter compression?

\section{Discussion}

In this paper, we demonstrate how spatial symmetries can be used to reduce the number of model parameters by a factor of 16 while improving its performance on test data. We show these results for 2-dimensional data, though it can be extended to 3 dimensions using the same logic, or applied to more modern variants such as the Multigrid Tensorized Fourier Neural Operator~\cite{kossaifi2023multi} or the Spherical Fourier Neural Operators~\cite{bonev2023spherical}, which are often used for geophysical or weather modelling, though with more difficulty in the relation between the Spherical Harmonic modes. Here we have demonstrated how symmetries can be used when applied to scalar field inputs. Many problems involve vector or pseudo-vector inputs or outputs, like fluid velocity or vorticity respectively. The ideas presented in this model could be extended to work with different types of fields which may have different symmetry requirements. Alternatively, some models use time as a third dimension to the input, $u(x, y, t)$ which again requires different constraints such as generating the Fourier kernel from symmetries in $x$ and $y$ but not $t$. A more general model could be developed that incorporates different symmetries requirements for each channel and generate the appropriate constraints dynamically.  Overall, these ideas demonstrate how symmetry can be incorporated into Neural Operators to reduce the model parameters while improving performance, and can be adapted into more complex and efficient formulations of the Fourier Neural Operators that have been developed recently~\cite{duruisseaux2025fourier} to further boost performance and reduce model size.

\bibliography{refs}% Produces the bibliography via BibTeX.

@article{li2020fourier,
  title={Fourier neural operator for parametric partial differential equations},
  author={Li, Zongyi and Kovachki, Nikola and Azizzadenesheli, Kamyar and Liu, Burigede and Bhattacharya, Kaushik and Stuart, Andrew and Anandkumar, Anima},
  journal={arXiv preprint arXiv:2010.08895},
  year={2020}
}

@article{helwig2023group,
  title={Group equivariant fourier neural operators for partial differential equations},
  author={Helwig, Jacob and Zhang, Xuan and Fu, Cong and Kurtin, Jerry and Wojtowytsch, Stephan and Ji, Shuiwang},
  journal={arXiv preprint arXiv:2306.05697},
  year={2023}
}

@article{tran2021factorized,
  title={Factorized fourier neural operators},
  author={Tran, Alasdair and Mathews, Alexander and Xie, Lexing and Ong, Cheng Soon},
  journal={arXiv preprint arXiv:2111.13802},
  year={2021}
}

@article{li2025d,
  title={D-FNO: A decomposed Fourier neural operator for large-scale parametric partial differential equations},
  author={Li, Kangjie and Ye, Wenjing},
  journal={Computer Methods in Applied Mechanics and Engineering},
  volume={436},
  pages={117732},
  year={2025},
  publisher={Elsevier}
}

@article{kossaifi2023multi,
  title={Multi-grid tensorized fourier neural operator for high-resolution pdes},
  author={Kossaifi, Jean and Kovachki, Nikola and Azizzadenesheli, Kamyar and Anandkumar, Anima},
  journal={arXiv preprint arXiv:2310.00120},
  year={2023}
}

@article{shen2021rotation,
  title={Rotation equivariant operators for machine learning on scalar and vector fields},
  author={Shen, Paul and Herbst, Michael and Viswanathan, Venkat},
  journal={arXiv preprint arXiv:2108.09541},
  year={2021}
}

@book{temam2024navier,
  title={Navier--Stokes equations: theory and numerical analysis},
  author={Temam, Roger},
  volume={343},
  year={2024},
  publisher={American Mathematical Society}
}

@article{turing1990chemical,
  title={The chemical basis of morphogenesis},
  author={Turing, Alan Mathison},
  journal={Bulletin of mathematical biology},
  volume={52},
  number={1},
  pages={153--197},
  year={1990},
  publisher={Springer}
}

@article{lighthill1955kinematic,
  title={On kinematic waves II. A theory of traffic flow on long crowded roads},
  author={Lighthill, Michael James and Whitham, Gerald Beresford},
  journal={Proceedings of the royal society of london. series a. mathematical and physical sciences},
  volume={229},
  number={1178},
  pages={317--345},
  year={1955},
  publisher={The Royal Society London}
}

@article{richards1956shock,
  title={Shock waves on the highway},
  author={Richards, Paul I},
  journal={Operations research},
  volume={4},
  number={1},
  pages={42--51},
  year={1956},
  publisher={INFORMS}
}

@article{fefferman2006existence,
  title={Existence and smoothness of the Navier-Stokes equation},
  author={Fefferman, Charles L},
  journal={The millennium prize problems},
  volume={57},
  number={67},
  pages={22},
  year={2006}
}

@article{black1973pricing,
  title={The pricing of options and corporate liabilities},
  author={Black, Fischer and Scholes, Myron},
  journal={Journal of political economy},
  volume={81},
  number={3},
  pages={637--654},
  year={1973},
  publisher={The University of Chicago Press}
}

@article{merton1971theory,
  title={Theory of rational option pricing},
  author={Merton, Robert C and others},
  year={1973},
  journal={The Bell Journal of Economics and Management Science},
  publisher={The RAND Corporation}
}

@article{duruisseaux2025fourier,
  title={Fourier Neural Operators Explained: A Practical Perspective},
  author={Duruisseaux, Valentin and Kossaifi, Jean and Anandkumar, Anima},
  journal={arXiv preprint arXiv:2512.01421},
  year={2025}
}

@inproceedings{bonev2023spherical,
  title={Spherical fourier neural operators: Learning stable dynamics on the sphere},
  author={Bonev, Boris and Kurth, Thorsten and Hundt, Christian and Pathak, Jaideep and Baust, Maximilian and Kashinath, Karthik and Anandkumar, Anima},
  booktitle={International conference on machine learning},
  pages={2806--2823},
  year={2023},
  organization={PMLR}
}

@book{lapidus1999numerical,
  title={Numerical solution of partial differential equations in science and engineering},
  author={Lapidus, Leon and Pinder, George F},
  year={1999},
  publisher={John Wiley \& Sons}
}

@book{kalnay2003atmospheric,
  title={Atmospheric modeling, data assimilation and predictability},
  author={Kalnay, Eugenia},
  year={2003},
  publisher={Cambridge university press}
}

@article{dai2023neural,
  title={Neural operator learning for ultrasound tomography inversion},
  author={Dai, Haocheng and Penwarden, Michael and Kirby, Robert M and Joshi, Sarang},
  journal={arXiv preprint arXiv:2304.03297},
  year={2023}
}

@article{gopakumar2024plasma,
  title={Plasma surrogate modelling using Fourier neural operators},
  author={Gopakumar, Vignesh and Pamela, Stanislas and Zanisi, Lorenzo and Li, Zongyi and Gray, Ander and Brennand, Daniel and Bhatia, Nitesh and Stathopoulos, Gregory and Kusner, Matt and Peter Deisenroth, Marc and others},
  journal={Nuclear Fusion},
  volume={64},
  number={5},
  pages={056025},
  year={2024},
  publisher={IOP Publishing}
}

@article{watt2025ace2,
  title={ACE2: accurately learning subseasonal to decadal atmospheric variability and forced responses},
  author={Watt-Meyer, Oliver and Henn, Brian and McGibbon, Jeremy and Clark, Spencer K and Kwa, Anna and Perkins, W Andre and Wu, Elynn and Harris, Lucas and Bretherton, Christopher S},
  journal={npj Climate and Atmospheric Science},
  volume={8},
  number={1},
  pages={205},
  year={2025},
  publisher={Nature Publishing Group UK London}
}

@article{li2024physics,
  title={Physics-informed neural operator for learning partial differential equations},
  author={Li, Zongyi and Zheng, Hongkai and Kovachki, Nikola and Jin, David and Chen, Haoxuan and Liu, Burigede and Azizzadenesheli, Kamyar and Anandkumar, Anima},
  journal={ACM/IMS Journal of Data Science},
  volume={1},
  number={3},
  pages={1--27},
  year={2024},
  publisher={ACM New York, NY}
}

@article{cai2021physics,
  title={Physics-informed neural networks (PINNs) for fluid mechanics: A review},
  author={Cai, Shengze and Mao, Zhiping and Wang, Zhicheng and Yin, Minglang and Karniadakis, George Em},
  journal={Acta Mechanica Sinica},
  volume={37},
  number={12},
  pages={1727--1738},
  year={2021},
  publisher={Springer}
}

@article{white2023stabilized,
  title={Stabilized neural differential equations for learning dynamics with explicit constraints},
  author={White, Alistair and Kilbertus, Niki and Gelbrecht, Maximilian and Boers, Niklas},
  journal={Advances in Neural Information Processing Systems},
  volume={36},
  pages={12929--12950},
  year={2023}
}

@article{raissi2019deep,
  title={Deep learning of vortex-induced vibrations},
  author={Raissi, Maziar and Wang, Zhicheng and Triantafyllou, Michael S and Karniadakis, George Em},
  journal={Journal of fluid mechanics},
  volume={861},
  pages={119--137},
  year={2019},
  publisher={Cambridge University Press}
}

\end{document}